%% file: Writing.tex
\documentclass[conference]{IEEEtran}
\IEEEoverridecommandlockouts
\usepackage{cite}
\usepackage{amsmath,amssymb,amsfonts}
\usepackage{algorithmic}
\usepackage{graphicx}
\usepackage{textcomp}
\usepackage{xcolor}
\def\BibTeX{{\rm B\kern-.05em{\sc i\kern-.025em b}\kern-.08em
    T\kern-.1667em\lower.7ex\hbox{E}\kern-.125emX}}
\usepackage{ifpdf}
\usepackage{pgfplots}
\usepackage{tikz}
\usepackage{subfig}                     % for subfigures
\usepackage{float}
\usepackage{booktabs}                   % nice looking tables (for tables with ONLY horizontal lines)

\usepackage[T1]{fontenc}
\usepackage[utf8]{inputenc}
\usepackage[utf8]{inputenc}
\usepackage{pgfplots}
\usepackage{grffile}
\pgfplotsset{compat=newest}
\usetikzlibrary{plotmarks}
\usetikzlibrary{arrows.meta}
\usepgfplotslibrary{patchplots}
\usepackage{subfig}
\usepackage{siunitx}
\usepackage{bmpsize}
\usepackage[abs]{overpic}
\usepackage{adjustbox}
\usepackage{booktabs,array}
\usepackage{multirow}
\usepackage{balance}
\renewcommand{\thetable}{\Roman{table}}
\def\tablename{TABLE}
\usepackage{units}

\DeclareRobustCommand*{\IEEEauthorrefmark}[1]{%
	\raisebox{0pt}[0pt][0pt]{\textsuperscript{\footnotesize #1}}%
}

\makeatletter
\renewcommand\footnotesize{%
	\@setfontsize\footnotesize\@ixpt{9.5}%
	\abovedisplayskip 8\p@ \@plus2\p@ \@minus4\p@
	\abovedisplayshortskip \z@ \@plus\p@
	\belowdisplayshortskip 4\p@ \@plus2\p@ \@minus2\p@
	\def\@listi{\leftmargin\leftmargini
		\topsep 4\p@ \@plus2\p@ \@minus2\p@
		\parsep 2\p@ \@plus\p@ \@minus\p@
		\itemsep \parsep}%
	\belowdisplayskip \abovedisplayskip
}
\makeatother

\renewcommand{\baselinestretch}{0.93}
\begin{document}

\title{An Adversarial Super-Resolution Remedy for\\ Radar Design Trade-offs}
	\author{\IEEEauthorblockN{Karim Armanious\IEEEauthorrefmark{1}*,
			Sherif Abdulatif\IEEEauthorrefmark{1}*,
		Fady Aziz\IEEEauthorrefmark{2},
		Urs Schneider\IEEEauthorrefmark{2} and
		Bin Yang\IEEEauthorrefmark{1}}
	\IEEEauthorblockA{\IEEEauthorrefmark{1}Institute of Signal Processing and System Theory, University of Stuttgart, Germany\\
		\IEEEauthorrefmark{2}Fraunhofer Institute for Manufacturing Engineering and Automation IPA, Stuttgart, Germany}
	Email: \{karim.armanious, sherif.abdulatif\}@iss.uni-stuttgart.de\\
	*These authors contributed to this work equally.}

\maketitle

\begin{abstract}

Radar is of vital importance in many fields,
such as autonomous driving, safety and surveillance applications.
However, it suffers from stringent constraints on its design 
parametrization leading to multiple trade-offs.
For example, the bandwidth in FMCW radars is inversely proportional with
both the maximum unambiguous range and range resolution. In this work, 
we introduce a new method for circumventing radar design trade-offs.
We propose the use of recent advances in computer vision, more specifically
generative adversarial networks (GANs), to enhance low-resolution radar 
acquisitions into higher resolution counterparts while maintaining the 
advantages of the low-resolution parametrization. The capability of the 
proposed method was evaluated on the velocity resolution and range-azimuth trade-offs in micro-Doppler 
signatures and FMCW uniform linear array (ULA) radars, respectively.
\end{abstract}

\vspace{-2mm}
\section{Introduction}
Radar is one of the most powerful tools for environment sensing. This is due to its superior capabilities in harsh environments and under low lighting conditions, where cameras and other vision-based sensors usually fail to operate. However, the design of a radar system for a particular application is often subject to some practical constraints and trade-offs between different system parameters \cite{murad13}. For instance, the bandwidth is inversely proportional to both the range resolution and the maximum unambiguous detectable range. Thus, having a higher range resolution will lead to a deterioration in the maximum unambiguous detectable range. The same applies for the velocity. Faster chirps will enhance the velocity resolution at the expense of a lower maximum detectable velocity \cite{richards2005fundamentals}. For MIMO radars, increasing the number of antennas will enhance the angular resolution but the longer acquisition time impairs real-time applications \cite{geibig16}. 

Accordingly, radar systems experience a fundamental limitation in many applications and they can not be optimum in all properties simultaneously. As a result, radar is often used in fusion with other sensors and not as a standalone sensor. For example, autonomous driving is the most actively researched area where radar is utilized in addition to other electro-optical sensors such as camera and LIDAR \cite{dolgov2016modifying}. These sensors can either operate independently or co-dependently to construct 3D models of the surroundings. Recently, radar is also employed for indoor monitoring applications due to its insensitivity against smoke and lighting conditions. For instance, a fusion between radar and thermal infrared camera (TIR) is proposed in \cite{ulrich18} to detect real hostages from mirrored TIR reflections in a fire-fighting scenario.

Additionally, radar imaging applications are also susceptible to radar design trade-offs. In these applications, a scene is constructed as a 2D or 3D image using either electronic beam steering \cite{geibig16} or mechanical scanning, such as synthetic aperture radar (SAR) \cite{lasaponara2013satellite} or inverse synthetic aperture radar (ISAR) \cite{ozdemir2012inverse}. Active research in radar imaging leads to the evolution of different areas and applications such as through wall imaging \cite{amin2017through}, land and weather monitoring \cite{carpenter2001parametric}. However, these techniques suffer from trade-offs between their spatial parameters (range-azimuth-elevation resolutions) due to design constraints on bandwidth, acquisition time, number of antennas and antenna separation.

Several methods have been proposed to enhance the acquired radar measurements. This family of methods is called radar super-resolution. They operate in general on the collected IQ time-series. In \cite{torres2007initial}, the authors used finer sampling to enhance resolution of both range and azimuth in NEXRAD weather data. They also utilized SZ-Phase coding on the time series data to enhance the Doppler resolution on the same dataset \cite{hubbert2005}. Super-resolution of SAR and ISAR imaging is also extensively studied by treating the problem as sparse signal reconstruction based on an orthogonal basis. Then convex optimization or Bayesian deconvolution are used to solve the reconstruction problem \cite{xu11,samadi2009}. A very similar approach is applied in \cite{xia13} to enhance the resolution of direction of arrival (DOA) estimation using MUSIC algorithm.

In this work, we propose a new technique for radar super-resolution to bypass the aforementioned practical constraints and trade-offs regarding radar parametrization. The proposed technique operates on the post-processed 2D time-frequency or range-azimuth representations/images. Unlike the previous approaches operating on time-series IQ data, this more accessible image-based approach allows the utilization of state-of-the-art techniques from the deep learning community to achieve radar super-resolution.

In recent years, the significant increase in data size and computational resources leads to a considerable boost in deep learning. Especially the development of deep convolutional neural networks (DCNNs) radically changed the research approach in the field of image processing. Generative adversarial networks (GANs), introduced by Ian Goodfellow in \cite{ian2014GAN}, are a recent branch and are considered as the state-of-the-art in image generation tasks. Moreover, a variant of GANs, conditional generative adversarial networks (cGANs) \cite{isola2016L1}, was recently introduced for image-to-image translation such as natural image super-resolution \cite{ledig2017photo}, medical image correction and inpainting \cite{27,277}.

In this paper, we utilize a modified cGAN framework for the task of radar super-resolution. The proposed framework incorporates additional non-adversarial losses to enhance the details of the super-resolved radar images. To validate the effect of the proposed framework, experiments were conducted on the super-resolution of micro-Doppler ($\boldsymbol{\mu}$-D) signatures of walking human targets. Additionally, some preliminary experiments on the super-resolution of range-azimuth spectrum in FMCW uniform linear array (ULA) radar were conducted. 

\section{Experimental Setup}
\subsection{Micro-Doppler Signatures \label{sec:mD}}
The $\boldsymbol{\mu}$-D signature of a moving target is identified by the time-frequency (TF) analysis of the  backscattered IQ signal. The frequency shifts in the down-converted backscattered signal represent the velocity information of the specific target. It can be used to visualize the velocity variation of a certain target of interest over time. Accordingly, a moving rigid body, such as pendulum, will have only one velocity component which can be either varying or constant. On the other hand, complex bodies, such as humans, will have different velocity components representing the motion of each limb during a gait cycle \cite{microdopplerBook}.

In the conducted experiments, the proposed adversarial framework for radar super-resolution will be applied on the $\boldsymbol{\mu}$-D signatures of walking humans. The main motivation is that the $\boldsymbol{\mu}$-D signatures contain rich identifying information. Thus, they can be used to fingerprint a walking person based on unique walking style and body mass index (BMI) \cite{abdulatif2019}. As such, the proposed radar super-resolution framework aims at reconstructing a high-resolution $\boldsymbol{\mu}$-D signature from a low-resolution one while preserving the identity information of each person.
 
The TF representation is derived from the short-time Fourier transform (STFT) of the complex time-domain baseband signal. In our method, a Gaussian window with a fixed window size of 512 and a window overlap of 75\% is used to obtain all  $\boldsymbol{\mu}$-D signatures. A CW radar operating at a carrier frequency of $f_c =$ \unit[25]{GHz} is used for data collection. Based on the relations presented in \cite{microdopplerBook}, the maximum unambiguous velocity ($v_{max}$) and the corresponding velocity resolution ($v_{res}$) of the TF representation are
\begin{equation}
	v_{max}=\frac{c f_{p}}{4 f_{c}}\hspace{7.5mm} \text{and} \hspace{7.5mm}v_{res}=\frac{2 v_{max}}{w} \label{eq:vmax}
\end{equation} 
 where $c$ is the speed of light, $f_p$ is the pulse repetition frequency and $w$ is the STFT window size.
 
 A clear trade-off exists between the maximum unambiguous velocity range and the resultant velocity resolution. An enhancement of $v_{max}$ will lead to a corresponding deterioration of the resolution and vice versa. In this work, we aim to overcome this radar design trade-off by translating a low-resolution $\boldsymbol{\mu}$-D signature to a corresponding high-resolution signature while keeping the same window size and hence the STFT complexity.
 
   \begin{figure}
 	\vspace{-5mm}
 	\centering
 	\hspace{-4mm}\subfloat[Low resolution $\boldsymbol{\mu}$-D. \hspace{-5mm} ]{\resizebox{.82\columnwidth}{!}{\input{lRes.tex}}\label{fig:LD}}\\ \vspace{-2mm}
 	\hspace{-4mm}\subfloat[High resolution $\boldsymbol{\mu}$-D. \hspace{-5mm} ]{\resizebox{.82\columnwidth}{!}{\input{hRes.tex}}\label{fig:HD}}
 	\caption{An example of a paired high and low-resolution $\boldsymbol{\mu}$-D signatures of one full gait cycle for the same subject. \label{fig:exps}}
 	\vspace{-6mm}
 \end{figure}
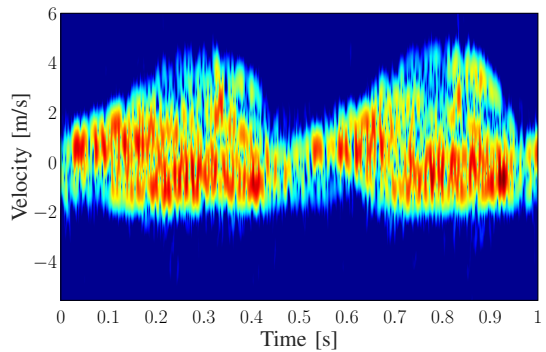
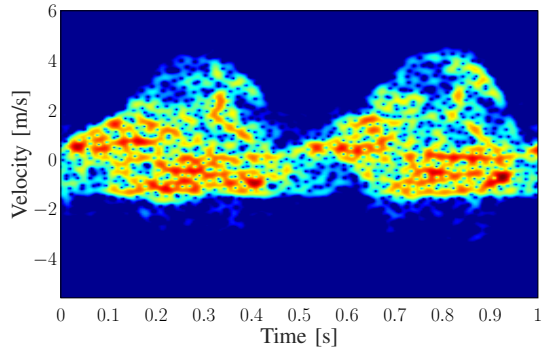
 \vspace{-2mm}
\subsection{Dataset Preparation and Collection}
In our experiments, we obtained low-resolution and high-resolution $\boldsymbol{\mu}$-D signatures by tuning the $f_p$ parameter based on Eq.~\ref{eq:vmax}. For the high-resolution case, $f_p$ is set to \unit[2]{kHz} to obtain a maximum velocity of \unit[6]{m/s} and a velocity resolution of \unit[2]{cm/s}. In the low-resolution experiment, $f_p$ is increased to \unit[8]{kHz} and accordingly the maximum velocity is quadrupled to \unit[24]{m/s} with a lower velocity resolution of \unit[8]{cm/s}. The $f_p$ can be controlled in offline data processing by sampling the acquired continuous-wave IQ data of a single measurement. Thus, perfectly paired low and high-resolution  $\boldsymbol{\mu}$-D signatures can be acquired using a single radar acquisition, as illustrated in Fig.~\ref{fig:exps}. This pixel-wise pairing of the signatures is mandatory for training the utilized adversarial framework.
  
For the low-resolution images, the STFT operation results in a velocity range of -24 to \unit[24]{m/s} over the selected window size. Nonetheless, for the high-resolution case only a velocity range of -6 to \unit[6]{m/s} is achievable. However, it was found out that a velocity of $\pm$\unit[6]{m/s} is sufficient to cover the maximum velocity of the human gait signature, as illustrated in Fig. 1. Thus, only a velocity range of -6 to \unit[6]{m/s} is considered for the low-resolution images resulting in an effective window size of 128 points compared to 512 for the high-resolution case. 

For dataset collection, 22 subjects of different genders, heights and weights are asked to walk on a treadmill at an intermediate velocity of \unit[1.5]{m/s}. The radar is placed behind the treadmill at a fixed distance of \unit[3]{m} and the direction of walking is always away from the radar. The experiment duration for each subject is \unit[180]{s} and then the collected IQ time-domain signals are decimated with different decimation factors to control the $f_p$ corresponding to both high and low-resolution signatures. Finally, the $\boldsymbol{\mu}$-D signatures of each walking subject is calculated as the STFT of the processed time signal as described above.

A full human gait cycle is divided into two swinging half gait cycles and each half gait is representing the limbs swinging velocity components for either left or right body side \cite{abdulatif2017real}. Since the targets are walking on a treadmill, the bulk motion of the targets is always around \unit[0]{m/s} and only the swinging limbs velocity components are preserved as depicted in Fig.~\ref{fig:exps}. Moreover, a clear difference between half swinging gait cycles of left and right body side is observed. Thus, the collected data is divided on half gait basis to add more variations in the dataset which will force the network to learn more detailed features. A total number of 5400 half gait signatures collected from 15 different subjects (360 cycles per subject) are used for training. The remaining unseen 7 subjects are used for testing the network performance over 2520 half gait signatures.
 
For training the adversarial framework, a final pre-processing step is required. The low and high-resolution signatures were resized to the same fixed size of 256$\times$256$\times$3 unsigned 16-bit images. This was achieved by using linear interpolation to up-scale the low-resolution signatures and down-scale the high-resolution counterparts to achieve a fixed image size. 

\section{Methodology}
The framework for radar spectrum super-resolution is based on cGANs with the inclusion of additional non-adversarial losses and a more advanced generator architecture. The proposed framework is illustrated in Fig.~\ref{fig:netArch}. 
\subsection{Conditional Generative Adversarial Networks \label{sec:cgan}}
The utilized framework consists of two Deep Convolutional Neural Networks (DCNNs), together forming a conditional adversarial framework \cite{isola2016L1}. The first network, the generator $G$, represents a mapping function from the input low-resolution $\boldsymbol{\mu}$-D signature $y$ to the corresponding high-resolution fake signature $\hat{x} = G(y)$. The second network, the discriminator $D$, receives as inputs the reconstructed high-resolution signature $\hat{x}$ together with the target ground truth signature $x$. It then proceeds to classify which is the real spectrum, $D(x,y) = 1$, and which is a fake spectrum, $D(\hat{x},y) = 0$. Both neural networks are trained against each other in competition, with the discriminator attempting to correctly classify the fake generated output and the generator attempting to improve the quality of the fake output to fool the discriminator. This training procedure is expressed as a min-max optimization task with respect to the generator and discriminator over the following adversarial loss function:
\begin{figure}
	\centering
	\includegraphics[width=0.49\textwidth]{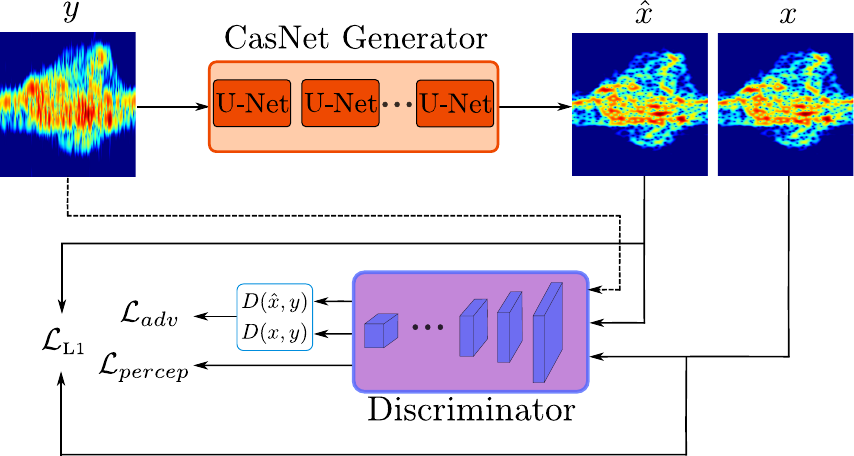}
	\caption{An overview of the proposed adversarial architecture for $\boldsymbol{\mu}$-D signatures super-resolution.\label{fig:netArch}}
	\vspace{-6mm}
\end{figure}
\begin{equation}
\min_{G} \max_{D} \mathcal{L}_{\small\textrm{adv}} = \mathbb{E}_{x,y} \left[\textrm{log} D(x,y) \right] + \mathbb{E}_{\hat{x},y} \left[\textrm{log} \left( 1 - D\left(\hat{x},y\right) \right) \right]
\end{equation}
\vspace{-8mm}
\begin{figure*}[t]
	\begin{minipage}[t]{1.0\linewidth}
		\centering
		\vspace{7mm}
		\begin{minipage}[t]{0.49\linewidth}
			\centering
			\begin{overpic}[width=0.31\textwidth]%
				{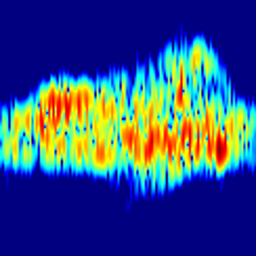}
				\centering
				\put(26,83){Input}
			\end{overpic}
			\begin{overpic}[width=0.31\textwidth]%
				{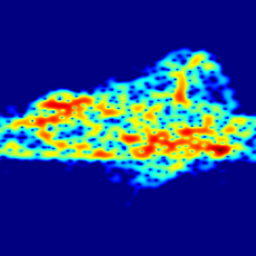}
				\centering
				\put(26,83){Output}
			\end{overpic}
			\begin{overpic}[width=0.31\textwidth]%
				{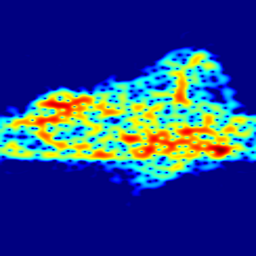}
				\centering
				\put(26,83){Target}
			\end{overpic}	
		\end{minipage}%
		\vline width 1pt
		\begin{minipage}[t]{0.49\linewidth}
			\centering
			\begin{overpic}[width=0.31\textwidth]%
				{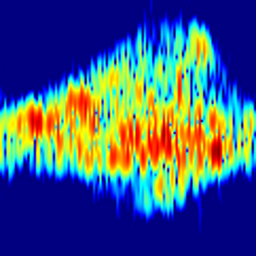}
				\centering
				\put(26,83){Input}
			\end{overpic}
			\begin{overpic}[width=0.31\textwidth]%
				{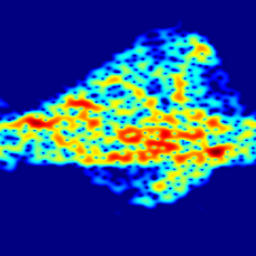}
				\centering
				\put(26,83){Output}
			\end{overpic}
			\begin{overpic}[width=0.31\textwidth]%
				{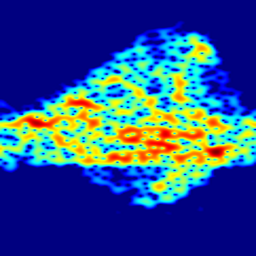}
				\centering
				\put(26,83){Target}
			\end{overpic}
		\end{minipage}\\
		\vspace{4mm}
		\begin{minipage}[t]{0.49\linewidth}
			\centering
			\begin{overpic}[width=0.31\textwidth]%
				{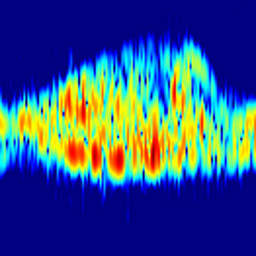}
			\end{overpic}
			\begin{overpic}[width=0.31\textwidth]%
				{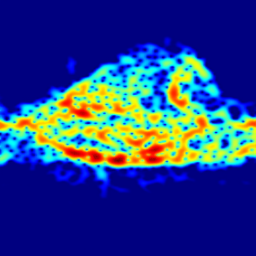}
			\end{overpic}
			\begin{overpic}[width=0.31\textwidth]%
				{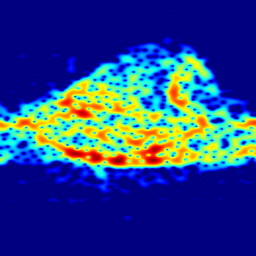}
			\end{overpic}	
		\end{minipage}%
		\vline width 1pt
		\begin{minipage}[t]{0.49\linewidth}
			\centering
			\begin{overpic}[width=0.31\textwidth]%
				{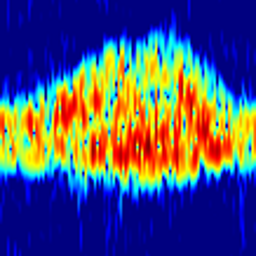}
			\end{overpic}
			\begin{overpic}[width=0.31\textwidth]%
				{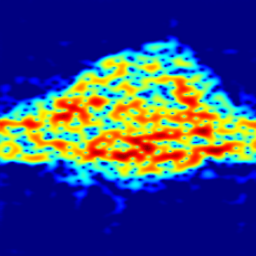}
			\end{overpic}
			\begin{overpic}[width=0.31\textwidth]%
				{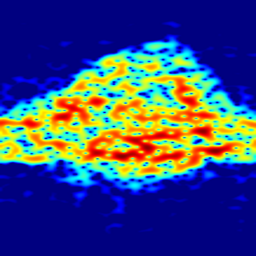}
			\end{overpic}
		\end{minipage}\\
	\end{minipage}
	\caption{Qualitative results for the super-resolution of human $\boldsymbol{\mu}$-D signature from a velocity resolution of \unit[8]{cm/s} to \unit[2]{cm/s}.}
	\label{5}
	\vspace{-4mm}
\end{figure*}
\subsection{Non-adversarial Losses}
It has been discussed previously that relying solely on the adversarial loss results in translated images with visual artifacts affecting the output quality. Thus, it has been suggested to incorporate auxiliary non-adversarial losses to further regularize the generator network. Inspired by the pix2pix framework, an additional L1 loss is utilized to penalize the pixel-wise discrepancy between the translated output and the ground truth target signatures \cite{isola2016L1}. This loss is given by
\begin{equation}
\mathcal{L}_{\small\textrm{L1}} = \mathbb{E}_{x,y} \left[\lVert{x - G(y)}\rVert_1\right]
\end{equation}
Moreover, to avoid blurring and inconsistencies in the local structures of the results, it has been suggested by previous adversarial frameworks, such as PAN and MedGAN \cite{wang18,armanious2018medgan}, to further utilize a feature-based perceptual loss to regularize the generator network. This loss function penalize the discrepancy between intermediate feature maps extracted by the discriminator network. The perceptual loss is given by
\begin{equation}
\mathcal{L}_{\small\textrm{Percep}} = \sum_{i = 0}^{L}  \lambda_{pi} \lVert{F_i\left(x\right) - F_i\left(\hat{x}\right)}\rVert_1
\end{equation}
where $F_i$ is the intermediate feature map extracted from the $i^{\textrm{th}}$ layer of the discriminator. $L$ is the total number of layers and $\lambda_{pi} > 0$ is the weight contribution for each individual layer. The final loss function utilized in the adversarial framework is the sum of the previously introduced adversarial, pixel and perceptual loss functions. 

\subsection{Architectural and Implementation Details}
Inspired by MedGAN, the super-resolution framework utilizes a CasNet architecture for the generator network \cite{armanious2018medgan}. CasNet concatenates in an end-to-end manner multiple U-Net architectures, introduced in \cite{ronneberger2015u}, to progressively refine the translated outputs. Each U-Net architecture consists of an encoder-decoder structure with multiple skip connections to avoid loss of information due to the bottleneck layer. The utilized CasNet architecture consists of a concatenation of five U-Nets architectures. For the discriminator architecture, a patch discriminator architecture, introduced by the pix2pix framework, was utilized \cite{isola2016L1}. The patch discriminator divides the input images into $70 \times 70$ patches before classifying each of them into being real or fake images. The classification score of all image patches is then averaged out. More in-depth architectural details and consideration can be found in \cite{armanious2018medgan}.

The framework was trained for 100 epochs using the ADAM optimizer \cite{kingma2014adam} on a single NVIDIA Titan X GPU. Training time was approximately 36 hours with an inference time per image of 150 ms during validation.

\section{Results}
To validate the capability of the utilized adversarial framework in the super-resolution $\boldsymbol{\mu}$-D signatures, the discussion will be based on qualitative and quantitative analysis. The qualitative results are presented in Fig. 3. Through learning the joint distribution mapping between the low-resolution signature to the corresponding high-resolution counterpart, the network is able to generalize to never seen signatures for new subjects not included in the training dataset. Furthermore, the results are not based on simple interpolation of the pixel-wise values. They are capable of maintaining both the global consistency as well as the local structure of the signature. Such preserved vital information represents identifying fingerprints which is unique in the signature of every subject.

Since the results show a high visual similarity between the output and the target signature, a quantitative comparison between both images is applied across the used test set. The correlation between the super-resolved $\boldsymbol{\mu}$-D signatures and their corresponding high-resolution ground truths was calculated to be 0.979 according to the structural similarity index (SSIM) \cite{wang2004ssim}, and 0.953 according to the universal quality index (UQI) \cite{sheikh2006image}. Moreover, a peak signal-to-noise ratio (PSNR) of \unit[22.83]{dB} was achieved. This quantitative result indicates that minimal vital information loss occurred during the super-resolution process.
\begin{figure*}
	\centering
	\includegraphics[width=0.96\textwidth]{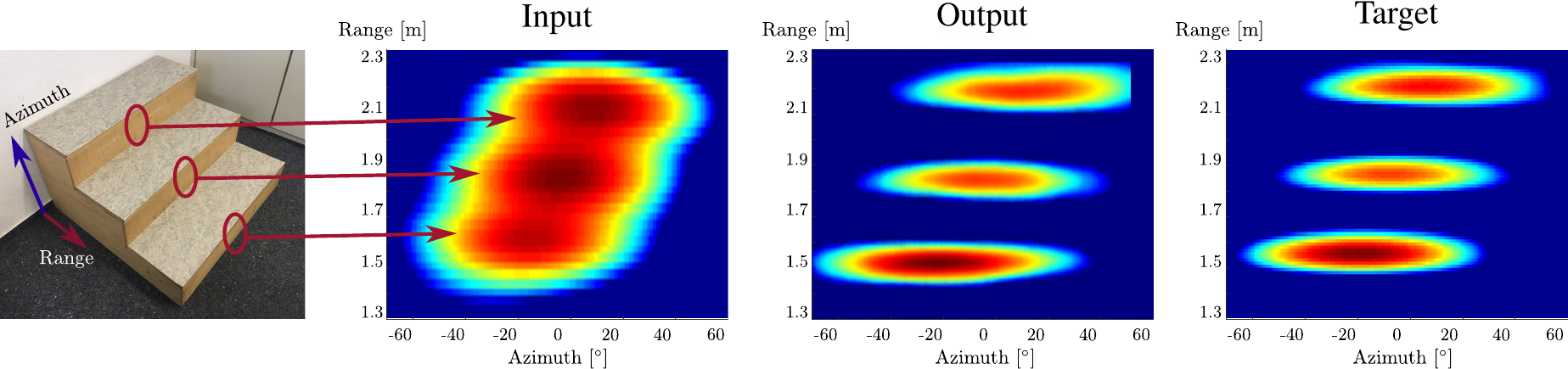}
	\caption{Results for radar super-resolution in the range-azimuth plane.\label{fig:netArch2}}
	\vspace{-6mm}
\end{figure*}
\section{ULA Use-Case and Discussion}
In this section, the proposed super-resolution framework was applied for the enhancement of the range-azimuth measurements of a ULA radar. During the parametrization of such a radar system, the range and the 3dB-angular resolutions are given by
\begin{equation}
R_{res}=\frac{c}{2 B}\hspace{4mm} \text{and} \hspace{4mm}\Psi{res}=0.89 \frac{\lambda}{N d \cos(\theta)} \label{eq:vmax2}
\end{equation}
where $B$ is the radar bandwidth, $N$ is the number of receiving elements, $d = \frac{\lambda}{2}$ is the fixed antenna separation, $\theta$ is the angle of arrival. As such, a ULA radar experiences multiple design trade-offs. For instance, increasing the bandwidth $B$ results in a better range resolution while having a detrimental effect on the maximum unambiguous range. Likewise, increasing the number of receiving elements enhances the radar angular resolution. However, it necessitates a longer acquisition time which may not be feasible in real-time applications \cite{richards2005fundamentals,geibig16}.

In this experiment, a dataset was collected over two separate acquisitions using a 77 GHz MIMO FMCW radar. This radar was operated effectively as a ULA with a single transmitting element. The radar was oriented such that the azimuth beam steering is in the $z$-plane and three-steps of a stair case were considered as the objects of interest in the measurements. The experiment setup and the scanning orientation of the radar are illustrated in Fig. 4. The two radar acquisitions were parametrized with $B = 1.2$ GHz and $N=4$ receiving elements for low-resolution acquisitions. For the high-resolution case, $B = 3.6$ GHz and $N = 8$ receiving elements. Thus, the range resolution is enhanced by a factor of three and the azimuth resolution by a factor of two. The proposed super-resolution framework was then trained to translate the low-resolution acquisitions into their corresponding high-resolution counterparts. A limited dataset of manually paired images were used for the training procedure.

The qualitative result of this preliminary experiment is shown in Fig. 4. It can be observed that lowering the utilized bandwidth and number of antennas deteriorates the resolution for both the range and azimuth such that the objects of interest can not be separated and are observed as a single entity. After the super-resolution translation by a cGAN, the range and azimuth resolutions are significantly enhanced and the three steps are clearly separated. The resulting range-azimuth spectrum looks almost the same as a radar with a parametrization of three-times the bandwidth and two-times the number of recieving elements. 

To summarize, in this work we propose the use of a cGAN framework for the super-resolution of different radar spectra. This is achievable since the cGAN framework is trained using paired datasets of low and high-resolution acquisitions. This enables the framework to learn the joint distribution mapping low-resolution data to that of higher resolutions as well as grasp the underlying structures of radar spectra. However, our work is not without limitations. Obtaining such paired datasets is often challenging in many practical applications, e.g. for MIMO or range-Doppler measurements with extensive planning and acquisition effort. Moreover, the trained cGAN framework is sensitive to the choice of the radar system and parametrization used within the utilized training dataset. 
By using data from different setups and applications to train the model, a certain generalization capability could be achieved.
%Thus, the model is not easily generalizable to new radar settings but will have to be re-trained for different use cases.
\section{Conclusion}

In this paper, a new technique for the super-resolution of radar signals was presented based on the utilization of a deep adversarial framework with additional non-adversarial losses. This technique aims to overcome the trade-offs in radar design parametrization by enhancing the post-processed 2D radar representations rather than operating on time-series IQ data.

To validate this proposition, experiments were carried out to enhance the velocity resolution, from \unit[8]{cm/s} to \unit[2]{cm/s}, for $\boldsymbol{\mu}$-D radar signatures while maximizing the unambiguous velocity range. Qualitative and quantitative analysis of the results indicate that the super-resolved $\boldsymbol{\mu}$-D signatures are highly correlated with their ground-truth high-resolution counterparts. Furthermore, the adversarial framework for super-resolution was applied in a preliminary experiment to enhance the resolution of range-azimuth measurements of a ULA radar with promising results.

In the future, we plan to examine the performance of the super-resolved $\boldsymbol{\mu}$-D signatures on post-processing tasks, such as object detection, localization and subject identification. We also aim to investigate the performance of adversarial super-resolution of ULA and MIMO radars more thoroughly by utilizing more test objects in training with a larger dataset. Additionally, extending the model to utilize recent unpaired adversarial translation techniques is crucial for extending the super-resolution framework for different radar measurements such as the range-Doppler maps in MIMO radars.  Finally, in the future we plan to compare the deep learning-based super-resolution approach with other classical approaches based on IQ time-series.

\bibliographystyle{IEEEtran}

%\bibliography{refs1}

\end{document}

%% file: lRes.tex
\begin{tikzpicture}

\begin{axis}[%
width=5.921in,
height=3.566in,
at={(0.758in,0.481in)},
scale only axis,
point meta min=-21.531091317488,
point meta max=18.468908682512,
axis on top,
xmin=0,
xmax=1,
xlabel style={font=\color{white!15!black}},
xlabel={\huge Time [s]},
ymin=-5.53824032,
ymax=6.00758272,
ytick style={draw=none},
xtick style={draw=none},
yticklabel style = {font=\fontsize{17}{0}},
xticklabel style = {font=\fontsize{17}{0}, yshift=-0.2cm},
ylabel style={font=\color{white!15!black}, yshift=-0.2cm},
ylabel={\huge Velocity [m/s]},
axis background/.style={fill=white},
legend style={legend cell align=left, align=left, draw=white!15!black}
]
\addplot [forget plot] graphics [xmin=0, xmax=1, ymin=-5.53824032, ymax=6.04306057070866] {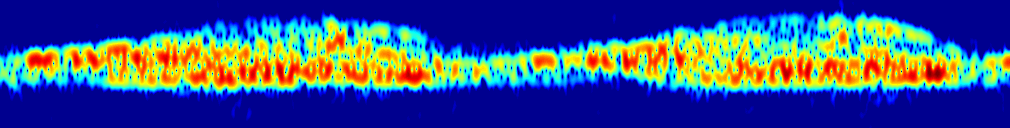};
\end{axis}

\begin{axis}[%
width=5.833in,
height=4.375in,
at={(0in,0in)},
scale only axis,
xmin=0,
xmax=1,
ymin=0,
ymax=1,
axis line style={draw=none},
ticks=none,
axis x line*=bottom,
axis y line*=left,
legend style={legend cell align=left, align=left, draw=white!15!black}
]
\end{axis}
\end{tikzpicture}%

%% file: hRes.tex
\begin{tikzpicture}

\begin{axis}[%
width=5.921in,
height=3.566in,
at={(0.758in,0.481in)},
scale only axis,
point meta min=-24.8423814227795,
point meta max=15.1576185772205,
axis on top,
xmin=0,
xmax=1,
xlabel style={font=\color{white!15!black}},
xlabel={\huge Time [s]},
ymin=-5.53824032,
ymax=6.00758272,
ytick style={draw=none},
xtick style={draw=none},
yticklabel style = {font=\fontsize{17}{0}},
xticklabel style = {font=\fontsize{17}{0}, yshift=-0.2cm},
ylabel style={font=\color{white!15!black}, yshift=-0.2cm},
ylabel={\huge Velocity [m/s]},
axis background/.style={fill=white},
legend style={legend cell align=left, align=left, draw=white!15!black}
]
\addplot [forget plot] graphics [xmin=0, xmax=1, ymin=-5.53824032, ymax=6.00758272] {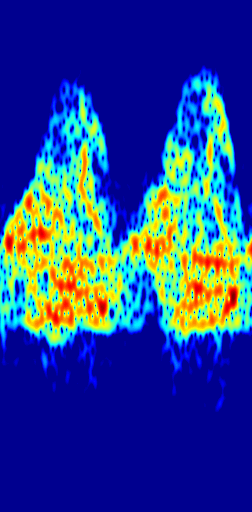};
\end{axis}

\begin{axis}[%
width=5.833in,
height=4.375in,
at={(0in,0in)},
scale only axis,
xmin=0,
xmax=1,
ymin=0,
ymax=1,
axis line style={draw=none},
ticks=none,
axis x line*=bottom,
axis y line*=left,
legend style={legend cell align=left, align=left, draw=white!15!black}
]
\end{axis}
\end{tikzpicture}%